\documentclass[11pt,a4paper]{article}
\usepackage[hyperref]{naaclhlt2018}
\usepackage{times}
\usepackage{latexsym}

\usepackage{url}

\aclfinalcopy 

\usepackage{amsmath}
\usepackage{graphicx}
\usepackage{tabulary}
\usepackage{makecell}
\usepackage{multirow}
\usepackage{csvsimple}
\usepackage{tabularx}
\usepackage{soul}
\usepackage{placeins}
\usepackage{subfigure}

\definecolor{greenC}{RGB}{128, 255, 0}
\definecolor{greenOk}{RGB}{0, 255, 191}
\definecolor{bad}{RGB}{255, 191, 0}

\title{A Mixed Hierarchical Attention based Encoder-Decoder Approach for Standard Table Summarization}

\author{Parag Jain$^\dagger$ \hspace{0.1cm} Anirban Laha$^{\dagger^*}$ \hspace{0.1cm} Karthik Sankaranarayanan$^{\dagger}$\\\bf Preksha Nema$^*$ \hspace{0.1cm} Mitesh M. Khapra$^{*\ddagger}$ \hspace{0.1cm} Shreyas Shetty$^*$\\
  $^\dagger$IBM Research \hspace{0.1cm} 
  $^*$IIT Madras, India\\
    $^{\ddagger}$ Robert Bosch Center for Data Science and Artificial Intelligence, IIT Madras \\
  {\tt \{pajain34,anirlaha,kartsank\}@in.ibm.com} \\ {\tt \{preksha,miteshk,shshett\}@cse.iitm.ac.in} \\}

\date{}

\hypersetup{draft}
\begin{document}
	\maketitle

\begin{abstract}
Structured data summarization involves generation of natural language summaries from structured input data. In this work, we consider summarizing structured data occurring in the form of tables as they are prevalent across a wide variety of domains. We formulate the \emph{standard table summarization} problem, which deals with tables conforming to a single predefined schema. To this end, we propose a \emph{mixed hierarchical attention} based encoder-decoder model which is able to leverage the structure in addition to the content of the tables. Our experiments on the publicly available \textsc{weathergov} dataset show around 18 BLEU ($\sim 30\%$) improvement over the current state-of-the-art.
\end{abstract}


\section{Introduction}

\begin{figure*}[ht]
\begin{center}
\includegraphics[width=1.0\linewidth]{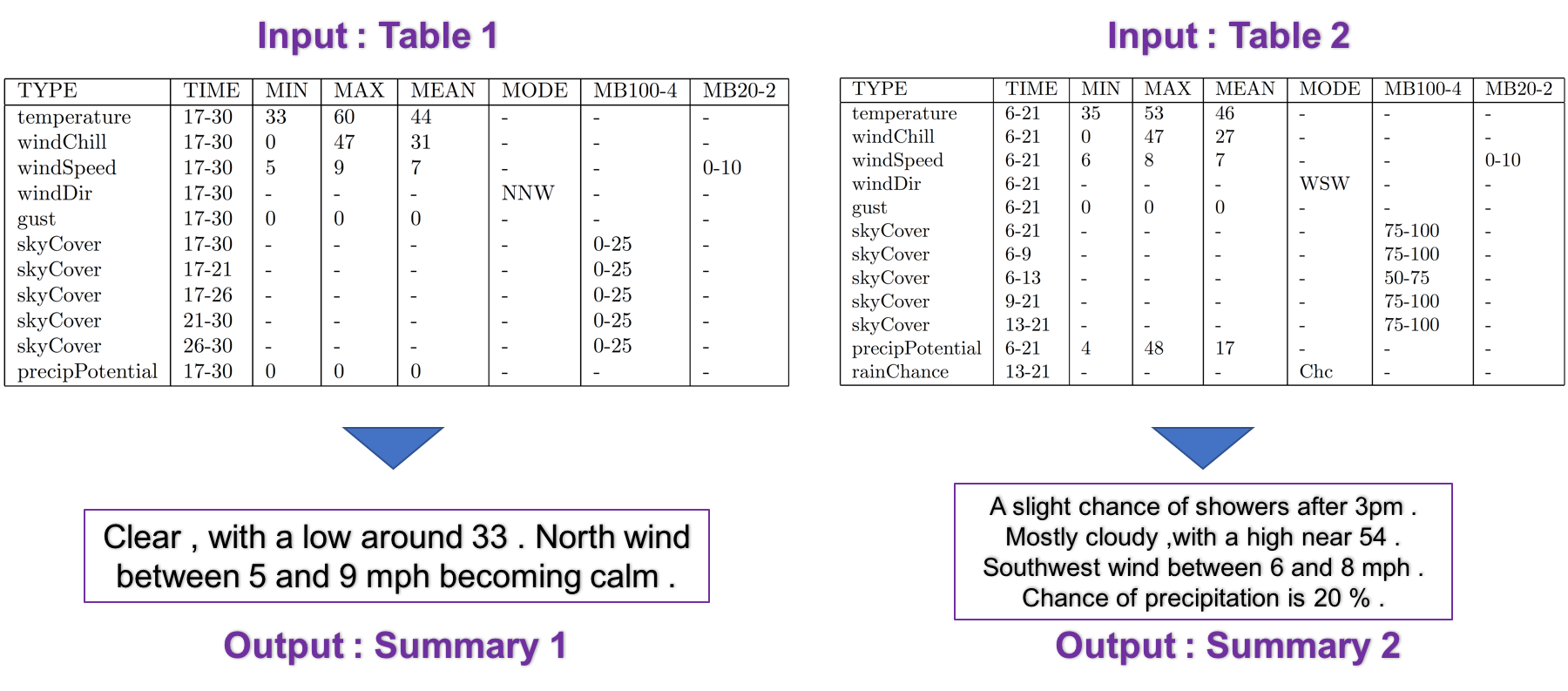}
\end{center}
\caption{Standard Table Summarization with fixed schema tables as input}
\label{fig:standard_table}
\end{figure*}

Abstractive summarization techniques from structured data seek to exploit both structure and content of the input data. The type of structure on the input side can be highly varied ranging from key-value pairs (e.g. \textsc{WikiBio} \cite{biography16}), source code \cite{source16}, ontologies \cite{owl14,dbpedia16}, or tables \cite{wiseman2017challenges}, each of which require significantly varying approaches. In this paper, we focus on generating summaries from tabular data. Now, in most practical applications such as finance, healthcare or weather, data in a table are arranged in rows and columns where the schema is known beforehand. However, change in the actual data values can necessitate drastically different output summaries. Examples shown in the figure \ref{fig:standard_table} have a predefined schema obtained from the \textsc{weathergov} dataset \cite{liang09} and its corresponding weather report summary. Therefore, the problem that we seek to address in this paper is to generate abstractive summaries of tables conforming to a predefined fixed schema (as opposed to cases where the schema is unknown). We refer to this setting as \textbf{standard table summarization} problem. Another problem that could be formulated is one in which the output summary is generated from multiple tables as proposed in a recent challenge \cite{wiseman2017challenges} (this setting is out of the scope of this paper). Now, as the schema is fixed, simple rule based  techniques \cite{Konstas2013InducingDP} or template based solutions could be employed. However, due to the vast space of selection (which attributes to use in the summary based on the current value it takes) and generation (how to express these selected attributes in natural language) choices possible, such approaches are not scalable in terms of the number of templates as they demand hand-crafted rules for both selection and generation.

We attempt to solve the problem of standard table summarization by leveraging the hierarchical nature of fixed-schema tables. In other words, rows consist of a fixed set of attributes and a table is defined by a set of rows. We cast this problem into a \textbf{mixed hierarchical attention model} following the encode-attend-decode \cite{attn15} paradigm. In this approach, there is static attention on the attributes to compute the row representation followed by dynamic attention on the rows, which is subsequently fed to the decoder. This formulation is theoretically more efficient than the fully dynamic hierarchical attention framework followed by \newcite{bowen16}. Also, our model does not need sophisticated sampling or sparsifying techniques like  \cite{LingR17,DengKLR17}, thus, retaining differentiability. To demonstrate the efficacy of our approach, we transform the publicly available \textsc{weathergov} dataset \cite{liang09} into fixed-schema tables, which is then used for our experiments. Our proposed \emph{mixed hierarchical attention model} provides an \textbf{improvement of around 18 BLEU (around 30\%) over the current state-of-the-art result} by \newcite{weather16}.  


	
\section{Tabular Data Summarization}
A standard table consist of set of records (or rows) $ R = (r_{1}, r_{2},...r_{T})$ and each record $r$ has a fixed set of attributes (or columns) $ A^{r} = (a_{r1},a_{r2},...a_{rM})$. Tables in figure \ref{fig:standard_table} have 7 columns (apart from `TYPE') which correspond to different attributes. Also $U = (u_1, u_2,...u_T)$ represents the type of each record where $u_k$ is one-hot encoding for the record type for record $r_k$. Training data consists of instance pairs $(X_i, Y_i)$ for $i = 1,2,..n$, where $X_i=(R_i, U_i)$ represents the input table and $Y_i=(y_1,...,y_{T'})$ represents the corresponding natural language summary. In this paper, we propose an end-to-end model which takes in a table instance $X$ to produce the output summary $Y$. This can be derived by solving in ${Y}$ the following conditional probability objective:
\begin{align}
\label{eq0}
Y^*  &= \arg\max_{Y} \prod_{t=1}^{T'} p(y_t| y_1, ..., y_{t-1}, X)
\end{align}

\begin{figure}[th] 
\centering
  \includegraphics[width=0.9\linewidth,keepaspectratio]{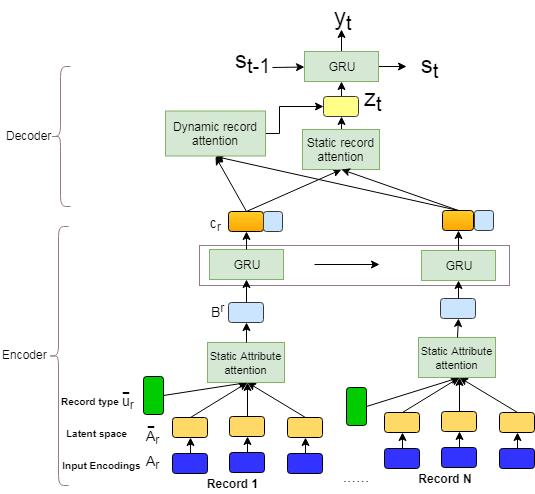}
  \caption{Proposed Architecture}
  \label{fig:model_arch}
\end{figure}

\subsection{Mixed Hierarchical Attention Model (MHAM)}
Our model is based on the \textit{encode-attend-decode} paradigm as defined by \newcite{attn15}. It consists of an \textit{encoder} RNN  which encodes a variable length input sequence $\textbf{x} = (x_1,...,x_T)$ into a representation sequence $\textbf{c}=(c_1,...,c_T)$. Another decoder RNN generates sequence of output symbols $\textbf{y} = (y_1,...,y_{T'})$, attending to different combinations of $c_i$ while generating different $y_t$.

As illustrated in figure \ref{fig:model_arch}, our encoder is not a single RNN. The encoder has a hierarchical structure to leverage the structural aspect of a \emph{standard table}: a table consists of a set of records (or rows) and each record consists of values corresponding to a fixed set of attributes. We call it a \emph{mixed hierarchical attention} based encoder, as it incorporates static attention and dynamic attention at two different levels of the encoder. At the record level, the attention over record representations is \emph{dynamic} as it changes with each decoder time step.  Whereas at the attribute level, since the schema is fixed, a record representation can be computed without the need of varying attention over attributes - thus \emph{static} attention is used. For example, with respect to \textsc{weathergov} dataset, a \emph{temperature} record will always be defined by the attributes like min, max and mean irrespective of the decoder time step. So, attention over attributes can be static. On the other hand, while generating a word by the decoder, there can be a preference of focusing on a record type say, \emph{temperature}, over some other type say, \emph{windSpeed}. Thus, dynamic attention is used across records.

\begin{align}
\small
    \label{eqn2}
      \bar{A_{j}^{r}} = W_jA_{j}^{r},\ \forall j \in [1,M]  \\
\label{eqn3}
        I^{r}_{j} = \bar{A_{j}^{r}} \odot \bar{u_{r}} \quad,\quad \bar{u_r} = W_0 u_r  \\
\label{eqn4}
        \alpha_{j}^{r} = softmax_j(I^r) \\
\label{eqn5}
        B^r = \sum\limits_{j} \alpha_{j}^{r}  \bar{A_{j}^{r}}   \\
\label{eqn6}
        c_r = \left[h_r;B^r\right] \quad,\quad {h_r} = GRU(B^r)   \\
\label{eqn7}
        g_r = \sigma(q^Ttanh(Pc_r)) \\
\label{eqn8}
        \beta^r_t = v^{T} tanh(W_{s}s_{t-1} + W_c c_{r})   \\
\label{eqn9}
        w^r_{t} = softmax_r(\beta_t)  \\
\label{eqn10}
                z_t = \sum\limits_{r}\gamma^r_{t}c_r \quad,\quad  \gamma^r_{t} = \frac{g_rw^r_{t}}{\sum\limits_{r}g_rw^r_{t}}  \\
\label{eqn11}
        s_t = GRU(z_t, s_{t-1})  \\
\label{eqn12}
        l_t = W_{1}s_{t} + W_{2}z_{t} + b_l  \\
\label{eqn13}
        p_t = softmax(l_t) 
\end{align}

	
\paragraph*{Capturing attribute semantics:}
We learn record type embeddings and use them to calculate attentions over attributes. \textit{For the trivial case of all records being same type, it boils down to having a single record type embedding.} Given attributes $A^{r}$ for a record $r$, where each attribute $a_{i}^{r}$ is encoded into a vector $A_{i}^{r}$ based on the attribute type (discussed further in section \ref{preprocessing}), using equation \ref{eqn2} we embed each attribute where $W_j$ is the embedding matrix for $j^{th}$ attribute. We embed record type one-hot vector $u_r$ through $W_0$, which is used to compute the importance score $I^r_j$ for attribute $j$ in record $r$ according to equation \ref{eqn3}.

\paragraph*{Static Attribute attention:}
	Not all attribute values contribute equally to the record. Hence, we introduce attention weights for attributes of each record. These attention weights are static and does not change with decoder time step. We calculate the attention probability vector $\alpha^{r}$ over attributes using the attribute importance vector $I^{r}$. The attention weights can then be used to calculate the record representation $B^r$ for record $r$ by using equations \ref{eqn4} and \ref{eqn5}.
	
	\paragraph*{Record Encoder:}
	A GRU based RNN encoder takes as input a sequence of attribute attended records $B^{1:N}$ and returns a sequence of hidden states $h_{1:N}$, where $h_{r}$ is the encoded vector for record $B^r$.
	We obtain the final record encoding $c_r$ (equation \ref{eqn6}) by concatenating the GRU hidden states with the embedded record encodings $B^r$.
	\paragraph*{Static Record attention:} 
	In a table, a subset of record types can always be more salient compared to other record types. This is captured by learning a static set of weights over all the records. These weights regulate the dynamic attention weights computed during decoding at each time step. Equation \ref{eqn7} performs this step where $g_r$ is the static record attention weight for $r^{th}$ record and $q$ and $P$ are weights to be learnt. We do not have any constraints on static attention vector. 
	
	\paragraph*{Dynamic Record attention for Decoder:}
	Our decoder is a GRU based decoder with dynamic attention mechanism similar to \cite{weather16} with modifications to modulate attention weights at each time step using static record attentions. At each time step $t$ attention weights are calculated using \ref{eqn8}, \ref{eqn9}, \ref{eqn10}, where $\gamma^r_{t}$ is the aggregated attention weight of record $r$ at time step $t$. We use the \textit{soft attention} over input encoder sequences $c_r$ to calculate the weighted average, which is passed to the GRU. GRU hidden state $s_t$ is used to calculate output probabilities $p_t$ by using a softmax as described by equation \ref{eqn11}, \ref{eqn12}, \ref{eqn13}, which is then used to get output word $y_t$.
	
Due to the static attention at attribute level, the time complexity of a single pass is $O(TM + TT')$, where $T$ is the number of records, $M$ is the number of attributes and $T'$ is the number of decoder steps. In case of dynamic attention at both levels (as in \newcite{bowen16}), the time complexity is much higher $O(TMT')$. Thus, \emph{mixed hierarchical attention} model is faster than fully dynamic hierarchical attention. For better understanding of the contribution of hierarchical attention(MHAM), we propose a simpler non-hierarchical (NHM) architecture with attention only at record level. In NHM, $B^r$ is calculated by concatenating all the record attributes along with corresponding record type.
	



\begin{table*} 
    \centering
    \begin{tabularx}{\textwidth}{|c|X|}
        \hline
    \textbf{\small Input Table} & \textbf{\small Generated Output} \\
      \hline
    \multirow{3}{*}{\raisebox{-.5\height}{\resizebox{0.33\textwidth}{!}{\begin{filecontents*}{images/ex2370.csv}
TYPE,TIME,MIN,MAX,MEAN,MODE,MB100-4,MB20-2
temperature,17-30,27,36,29,-,-,-
windChill,17-30,14,26,18,-,-,-
windSpeed,17-30,14,20,16,-,-,10-20
windDir,17-30,-,-,-,SSW,-,-
gust,17-30,0,0,0,-,-,-
skyCover,17-30,-,-,-,-,75-100,-
skyCover,17-21,-,-,-,-,75-100,-
skyCover,17-26,-,-,-,-,75-100,-
skyCover,21-30,-,-,-,-,75-100,-
skyCover,26-30,-,-,-,-,75-100,-
precipPotential,17-30,26,58,43,-,-,-
rainChance,17-21,-,-,-,Lkly,-,-
snowChance,17-30,-,-,-,Chc,-,-
snowChance,17-26,-,-,-,Lkly,-,-
\end{filecontents*}
\csvautotabular{images/ex2370.csv}}}} & \tiny  \textbf{Reference}: Periods of rain and possibly a thunderstorm . Some of the storms could produce heavy rain . Temperature rising to near 51 by 10am , then falling to around 44 during the remainder of the day . Breezy , with a north northwest wind between 10 and 20 mph . Chance of precipitation is 90 \% . New rainfall amounts between one and two inches possible . \\
    & \tiny \textbf{NHM}:Periods of rain and possibly a thunderstorm . Some of the storms could produce heavy rain . Temperature rising to near 51 by 8am , then falling to around 6 during the remainder of the day . Breezy , with a north northwest wind 10 to 15 mph increasing to between 20 and 25 mph . Chance of precipitation is 90 \% . New rainfall amounts between one and two inches possible . \\
    & \tiny \textbf{MHAM}: Periods of rain and possibly a \emph{thunderstorm} . Some of the storms could produce \emph{heavy rain} . Temperature rising to \emph{near 51 by 8am} , then falling to \emph{around 44} during the remainder of the day . Breezy , with a \emph{north northwest wind} between \emph{10 and 20 mph} . Chance of precipitation is \emph{90 \%} . New rainfall amounts between \emph{one and two inches} possible.\\
      \hline

      \multirow{3}{*}{\raisebox{-.5\height}{\resizebox{0.33\textwidth}{!}{\csvautotabular{images/ex2370.csv}}}} & \tiny \textbf{Reference}: A chance of rain and snow . Snow level 5500 feet . Mostly cloudy , with a low around 31 . Calm wind becoming \underline{north northeast} around 6 mph . Chance of precipitation is 40\% . \\
    & \tiny \textbf{NHM}: A chance of rain and snow . Mostly cloudy , with a low around 31 . North northwest wind at 6 mph becoming east southeast . Chance of precipitation is 40\% . \\
    & \tiny \textbf{MHAM}: A chance of \emph{rain and snow} . Snow level \emph{5800 feet lowering to 5300 feet} after midnight . Mostly cloudy , with a low \emph{around 31} . \emph{North northwest wind} at \emph{6 mph} becoming south southwest . Chance of precipitation is \emph{40\%} .\\
     &  \\ &  \\ \hline

      \multirow{3}{*}{\raisebox{-.5\height}{\resizebox{0.33\textwidth}{!}{\csvautotabular{images/ex2370.csv}}}} & \tiny \textbf{Reference}: Rain and snow likely , becoming all snow after 8pm . Cloudy , with a low around 22 . South southwest wind around 15 mph . Chance of precipitation is 60\% . New snow accumulation of less than one inch possible . \\
    & \tiny \textbf{NHM} : Rain or freezing rain likely before 8pm , then snow after 11pm , snow showers and sleet likely before 8pm , then a chance of rain or freezing rain after 3am . Mostly cloudy , with a low around 27 . South southeast wind between 15 and 17 mph . Chance of precipitation is 80\% . \underline{Little or no ice accumulation expected .} \underline{Little or no snow accumulation expected .} \\
    & \tiny \textbf{MHAM}: Snow , and freezing rain , \emph{snow after 9pm} . Cloudy , with a steady temperature \emph{around 23} . Breezy , with a \emph{south wind} between \emph{15 and 20 mph} . Chance of precipitation is \emph{60\%} . New snow accumulation of \emph{around an inch} possible .\\
      \hline
   
    \end{tabularx}
    \caption{ Anecdotal example. Records which contain all null attributes are not shown in the example. MB100-4 and MB20-2 correspond to mode-bucket-0-100-4 \& mode-bucket-0-20-2 resp. in the dataset.}
    \label{main_examples}
    \end{table*}
    
\section{Experiments}
\label{preprocessing}
\textbf{Dataset and methodology}: To evaluate our model we have used \textsc{weathergov} dataset \cite{liang09} which is the standard benchmark dataset to evaluate tabular data summarization techniques. We compared the performance of our model against the state-of-the-art work of MBW \cite{weather16}, as well as two other baseline models KL \cite{Konstas2013InducingDP} and ALK \cite{angeli2010simple}. Dataset consists of a total of 29,528 tables (25000:1000:3528 ratio for train:validation:test splits) corresponding to scenarios created by collecting weather forecasts for 3,753 cities in the U.S.A over three days. There are 12 record types consisting of both numeric and categorical values. Each table contains 36 weather records (e.g., temperature, wind direction etc.) along with a corresponding natural language summary.  
\paragraph*{Input Encodings:} Attributes were encoded based on the attribute type. Numbers are encoded in binary representation. Record type is encoded as a one-hot vector. Mode attribute is encoded using specific ordinal encodings for example `Lkly', `SChc', `Chc' are encoded as `00100000000000', `00010000000000' and `00001000000000' respectively. Similar works for directions, for example  `NW', `NNE' and `NE' are encoded as `00000100000000', `00000011000000' and `00000001000000' resp. Time interval were also encoded as ordinal encodings, for example `6-21' is encoded as `111100' and `6-13' is encoded as `110000', the six bits corresponding to six atomic time intervals available in the dataset. Other attributes and words were encoded as one-hot vectors.

\subsection{Training and hyperparameter tuning}
 We used TensorFlow \cite{tensorflow2015-whitepaper} for our experiments. Encoder embeddings were initialized by generating the values from a uniform distribution in the range [-1, 1). Other variables were initialized using Glorot uniform initialization \cite{Glorot10understandingthe}. We tune each hyperparameter by choosing parameter from a ranges of values, and selected the model with best sBLEU score in validation set over 500 epochs. We did not use any regularization while training the model. For both the models, the hyperparameter tuning was separately performed to give both models a fair chance of performance. For both the models, Adam optimizer \cite{kingma2014adam} was used with learning rate set to 0.0001. We found embedding size of 100, GRU size of 400, static record attention size$P$ of 150 to work best for MHAM model. We also experimented using bi-directional GRU in the encoder but there was no significant boost observed in the BLEU scores.


\textbf{Evaluation metrics}:
To evaluate our models we employed BLEU and Rouge-L scores. In addition to the standard BLEU (sBleu) \cite{papineni2002bleu}, a customized BLEU (cBleu) \cite{weather16} has also been reported. cBleu does not penalize numbers which differ by at most five; hence 20 and 18 will be considered same.

\section{Results and Analyses}
Table \ref{results} describes the results of our proposed models (MHAM and NHM) along with the aforementioned baseline models. We observe a significant performance improvement of 16.6 cBleu score (24\%) and 18.3 sBleu score (30\%) compared to the current state-of-the-art model of MBW. MHAM also shows an improvement over NHM in all metrics demonstrating the importance of hierarchical attention.
\begin{table}[h]
		\centering
		\begin{tabulary}{\linewidth}{LLCC}
			\hline
			\small Model                        & sBleu & cBleu  & Rouge-L \\ \hline
			\small KL & 36.54 & - & - \\
			\small ALK & 38.40 & 51.50 & - \\
			\small MBW & 61.0 & 70.4 & -  \\ 
			\hline
			\small NHM   & 76.2 & 85.0  & 86.4  \\ 
            \small \textbf{MHAM}               & \textbf{79.3} & \textbf{87.0}  & \textbf{88.5} \\ \hline
		\end{tabulary}
		\caption{Overall results}
		\label{results}
\end{table}

\begin{figure}
\centering     
\subfigure{\label{fig:a}\includegraphics[width=0.49\textwidth]{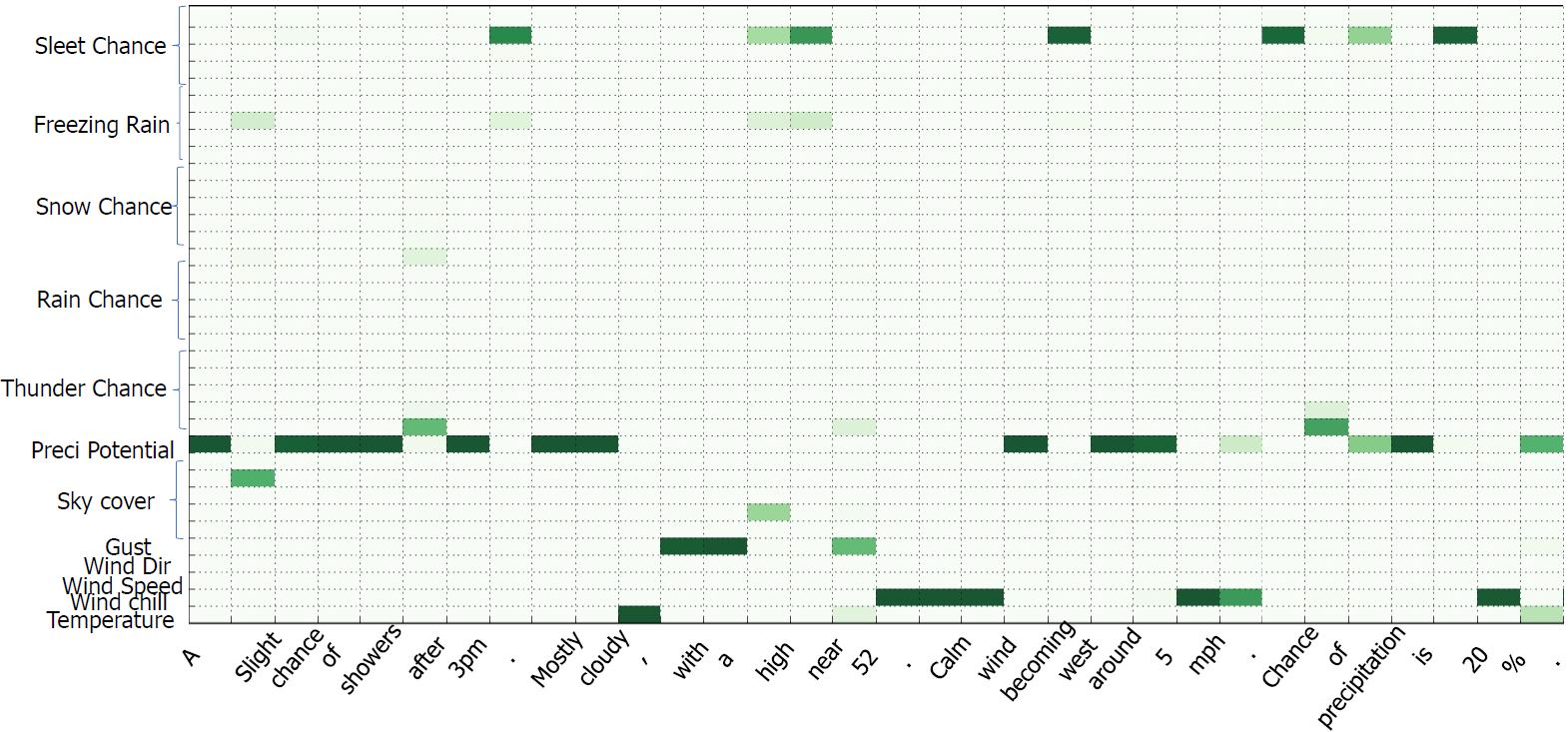}
}
\subfigure{\label{fig:b}\includegraphics[width=0.49\textwidth]{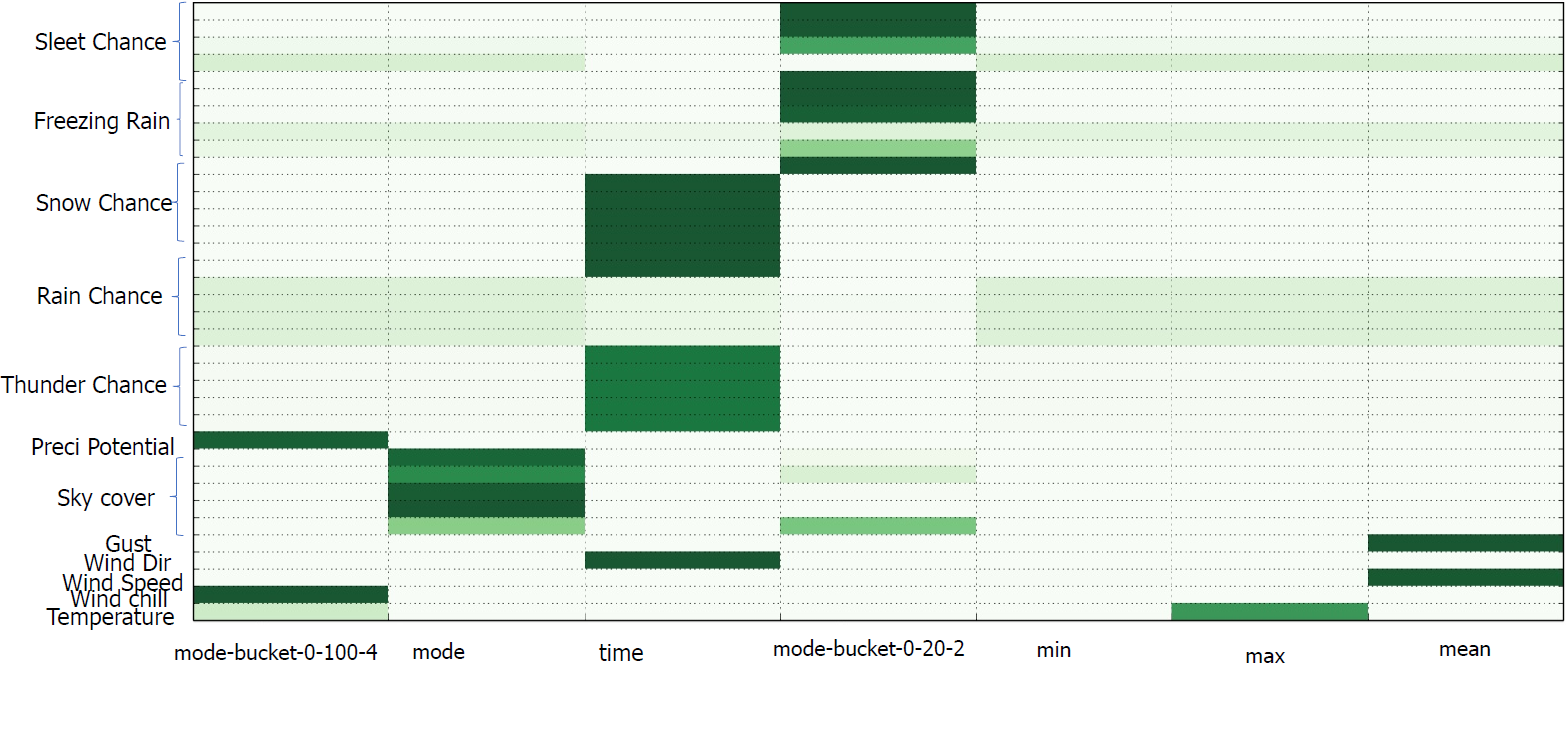}}
\caption{Heatmaps: (a) record level attention (top) and  (b) attribute level attention (bottom).}
\label{fig:heatmap}
\end{figure}

\textbf{Attention analysis:} Analysis of figure \ref{fig:heatmap} reveals that the learnt attention weights are reasonable. For example, as shown in figure \ref{fig:a}, for the phrase `with a high near 52', the model had a high attention on \emph{temperature} before and while generating the number `52'. Similarly while generating `mostly cloudy', the model had a high attention on \emph{precipitation potential}. Attribute attentions are also learned as expected (in figure \ref{fig:b}). The \emph{temperature}, \emph{wind speed} and \emph{gust} records have high weights on min/max/mean values which describe these records.

\textbf{Qualitative analysis:}
Table \ref{main_examples} contains example table-summary pairs, with summary  generated by the proposed hierarchical and non-hierarchical versions. We observe that our model is able to generate numbers more accurately by enabling hierarchical attention. Our model is also able to capture weak signals like \emph{snow accumulation}. Further, our proposed model MHAM is able to avoid repetition as compared to NHM. 

	
\section{Conclusion and Future Work}

In this work, we have formulated the problem of \emph{standard table summarization} where all the tables come from a predefined schema. Towards this, we proposed a novel mixed hierarchical attention based encoder-decoder approach. Our experiments on the publicly available \textsc{weathergov} benchmark dataset have shown significant improvements over the current state-of-the-art work.  Moreover, this proposed method is theoretically more efficient compared to the current fully dynamic hierarchical attention model. As future work, we propose to tackle general tabular summarization where the schema can vary across tables in the whole dataset.

\section{Acknowledgements}
We would like to acknowledge Abhijit Mishra for his valuable feedback and comments.
\FloatBarrier

\bibliography{naaclhlt2018}
\bibliographystyle{acl_natbib}
\end{document}


\appendix

\section{Supplementary Material}
\label{sec:supplemental}

\subsection{Dataset preprocessing}
\label{preprocessing}
We did not perform any case normalization over natural language summary text so that 'south' and 'South' are considered as two different words. Owing to this, the model in fact learned to generate words starting with upper case letters at the start of the sentence. All the words, numbers and punctuations are considered in their original form and tokenization was performed based on a single space character. 

\textbf{Standard Table:} The first step towards processing this dataset is transformation to a table with fixed-schema. If a certain attribute (say `mode') doesn't appear for a particular record (say `temperature'), even then that attribute is included as a column with `NULL' value for that record. Thus, all records will have fixed set of attributes, hence fixed schema.

\textbf{Input Encodings:} Each attribute in the table record is encoded based on the attribute type.. Numbers are encoded in binary representation. Record type is encoded as a one-hot vector. Mode attribute is encoded using specific ordinal encodings for example 'Lkly', 'SChc', 'Chc' are encoded as '00100000000000', '00010000000000' and '00001000000000' respectively. Similar works for directions, for example  'NW', 'NNE' and 'NE' are encoded as '00000100000000', '00000011000000' and '00000001000000' resp. Time interval were also encoded as ordinal encodings instead of one-hot vectors, for example '6-21' is encoded as '111100' and '6-13' is encoded as '110000', the six bits corresponding to six atomic time intervals available in the dataset. Other attributes and summary words were encoded as one-hot vectors.

\begin{table}[h]
\centering
  \begin{tabulary}{\linewidth}{LLCC}
\hline
      & No. of instances & Avg words/instance & Avg sent/instance \\ \hline
Train & 25000          & 30.2                & 3.27              \\
Test  & 3528           & 30.53               & 3.25              \\ 
Valid & 1000           & 31                  & 3.3        \\ \hline      
\end{tabulary}
\caption{Dataset statistics}\label{data-stats}
\end{table}

\subsection{Training and hyperparameter tuning}
The \textsc{weathergov} dataset is split into 25000 train samples, 1000 validation samples and 3528 samples for test set. Encoder embeddings were initialized by generating the values from a uniform distribution in the range [-1, 1). Other variables were initialized using Glorot uniform initialization \cite{Glorot10understandingthe}. We tune each hyperparameter by choosing parameter from a ranges of values, and selected the model with best sBLEU score in validation set. We did not use any regularization while training the model. For both the models, the hyperparameter tuning was separately performed to give both models a fair chance of performance. We used Adam optimizer \cite{kingma2014adam} for stochastic optimization on cross entropy loss for backpropagation. Table \ref{hyper} shows the hyperparameter setting which worked best for our proposed hierarchical (MHAM) and non-hierarchical models (NHM). For both the models, the learning rate(LR) was set to 0.0001. We also tried using bi-directional GRU in the encoder but there was no significant boost observed in the BLEU scores.

\begin{table}[h]
\centering
\small
\resizebox{\linewidth}{!}{%
\begin{tabularx}{\linewidth}{XXXXXXX}  
\hline
Model & GRU size & Enc emb & Record static attn.(P) & Record emb & Dec emb & LR \\ \hline
 MHAM  & 400                                                & 100                                                          & 150                                                                  & 150                                                                  & 250                                                              & 0.0001                                                  \\ 
 NHM   & 100                                                & 100                                                          & 200                                                                  & 200                                                                  & 200                                                              & 0.0001                                                  \\ \hline
  \end{tabularx}
  }
  \caption{Hyperparameter settings}
\label{hyper}

  \end{table}
  
\begin{table*} 
    \centering
    \begin{tabularx}{\textwidth}{|X|X|}
        \hline
    \textbf{Input table \& references} & \textbf{Generated summaries} \\
      \hline
      Example 1:&  \\
      \raisebox{-.5\height}{\resizebox{0.48\textwidth}{!}{\begin{filecontents*}{images/ex77.csv}
TYPE,TIME,MIN,MAX,MEAN,MODE,MB100-4,MB20-2
temperature,6-21,36,51,46,-,-,-
windChill,6-21,0,46,33,-,-,-
windSpeed,6-21,8,24,17,-,-,10-20
windDir,6-21,-,-,-,WNW,-,-
gust,6-21,0,34,6,-,-,-
precipPotential,6-21,30,90,75,-,-,-
thunderChance,13-21,-,-,-,Chc,-,-
rainChance,13-21,-,-,-,Def,-,-
\end{filecontents*}
\csvautotabular{images/ex77.csv}}} &  \textbf{NHM}:Periods of rain and possibly a thunderstorm . Some of the storms could produce heavy rain . Temperature rising to near 51 by 8am , then falling to around 6 during the remainder of the day . Breezy , with a north northwest wind 10 to 15 mph increasing to between 20 and 25 mph . Chance of precipitation is 90 \% . New rainfall amounts between one and two inches possible . \\
    \textbf{Reference}: Periods of rain and possibly a thunderstorm . Some of the storms could produce heavy rain . Temperature rising to near 51 by 10am , then falling to around 44 during the remainder of the day . Breezy , with a north northwest wind between 10 and 20 mph . Chance of precipitation is 90 \% . New rainfall amounts between one and two inches possible .    &  \textbf{MHAM}: Periods of rain and possibly a \emph{thunderstorm} . Some of the storms could produce \emph{heavy rain} . Temperature rising to \emph{near 51 by 8am} , then falling to \emph{around 44} during the remainder of the day . Breezy , with a \emph{north northwest wind} between \emph{10 and 20 mph} . Chance of precipitation is \emph{90 \%} . New rainfall amounts between \emph{one and two inches} possible. \\
    &  \\ \hline
     
     Example 2:&  \\
      \raisebox{-.5\height}{\resizebox{0.48\textwidth}{!}{\begin{filecontents*}{images/ex2842.csv}
TYPE,TIME,MIN,MAX,MEAN,MODE,MB100-4,MB20-2
temperature,17-30,32,42,34,-,-,-
windChill,17-30,0,0,0,-,-,-
windSpeed,17-30,2,7,4,-,-,0-10
windDir,17-30,-,-,-,NW,-,-
gust,17-30,0,0,0,-,-,-
precipPotential,17-30,37,48,38,-,-,-
rainChance,21-30,-,-,-,Chc,-,-
snowChance,26-30,-,-,-,Chc,-,-
\end{filecontents*}
\csvautotabular{images/ex2842.csv}}}  &  \textbf{NHM}: A chance of rain and snow . Mostly cloudy , with a low around 31 . North northwest wind at 6 mph becoming east southeast . Chance of precipitation is 40\% .\\ 
     \textbf{Reference}: A chance of rain and snow . Snow level 5500 feet . Mostly cloudy , with a low around 31 . Calm wind becoming \underline{north northeast} around 6 mph . Chance of precipitation is 40\% . &  \textbf{MHAM}: A chance of \emph{rain and snow} . Snow level \emph{5800 feet lowering to 5300 feet} after midnight . Mostly cloudy , with a low \emph{around 31} . \emph{North northwest wind} at \emph{6 mph} becoming south southwest . Chance of precipitation is \emph{40\%} .\\
     &  \\
      \hline
     Example 3:&  \\ 
     \raisebox{-.5\height}{\resizebox{0.48\textwidth}{!}{\csvautotabular{images/ex2370.csv}}} &  \textbf{NHM}: Rain or freezing rain likely before 8pm , then snow after 11pm , snow showers and sleet likely before 8pm , then a chance of rain or freezing rain after 3am . Mostly cloudy , with a low around 27 . South southeast wind between 15 and 17 mph . Chance of precipitation is 80\% . \underline{Little or no ice accumulation expected .} \underline{Little or no snow accumulation expected .}\\ 
     \textbf{Reference}: Rain and snow likely , becoming all snow after 8pm . Cloudy , with a low around 22 . South southwest wind around 15 mph . Chance of precipitation is 60\% . New snow accumulation of less than one inch possible . &  \textbf{MHAM}: Snow , and freezing rain , \emph{snow after 9pm} . Cloudy , with a steady temperature \emph{around 23} . Breezy , with a \emph{south wind} between \emph{15 and 20 mph} . Chance of precipitation is \emph{60\%} . New snow accumulation of \emph{around an inch} possible .\\
      &  \\
      \hline
   
    \end{tabularx}
    \caption{Anecdotal examples. In the interest of space, records which contain all null attributes are not shown in the examples. Attribute names MB100-4 and MB20-2 corresponds to mode-bucket-0-100-4 and mode-bucket-0-20-2 respectively in the dataset.}
    \label{examples}
    \end{table*}
    \FloatBarrier
    \bibliography{naaclhlt2018}
\bibliographystyle{acl_natbib}